\documentclass[conference]{IEEEtran}
\usepackage{cite}
\usepackage{amsmath,amssymb,amsfonts}
\usepackage{graphicx}
\usepackage{textcomp}
\usepackage{xcolor}
\def\BibTeX{{\rm B\kern-.05em{\sc i\kern-.025em b}\kern-.08em
    T\kern-.1667em\lower.7ex\hbox{E}\kern-.125emX}}

\usepackage{bm}
\usepackage{algorithm}
\usepackage{algorithmicx}
\usepackage{algpseudocode}
\usepackage{stfloats}
\usepackage{cuted}
\usepackage{capt-of}
\usepackage{hyperref}

\makeatletter
\let\@oldmaketitle\@maketitle
\renewcommand{\@maketitle}{\@oldmaketitle
   \begin{center}
      \setcounter{figure}{0}
      \includegraphics[width=0.9\linewidth]{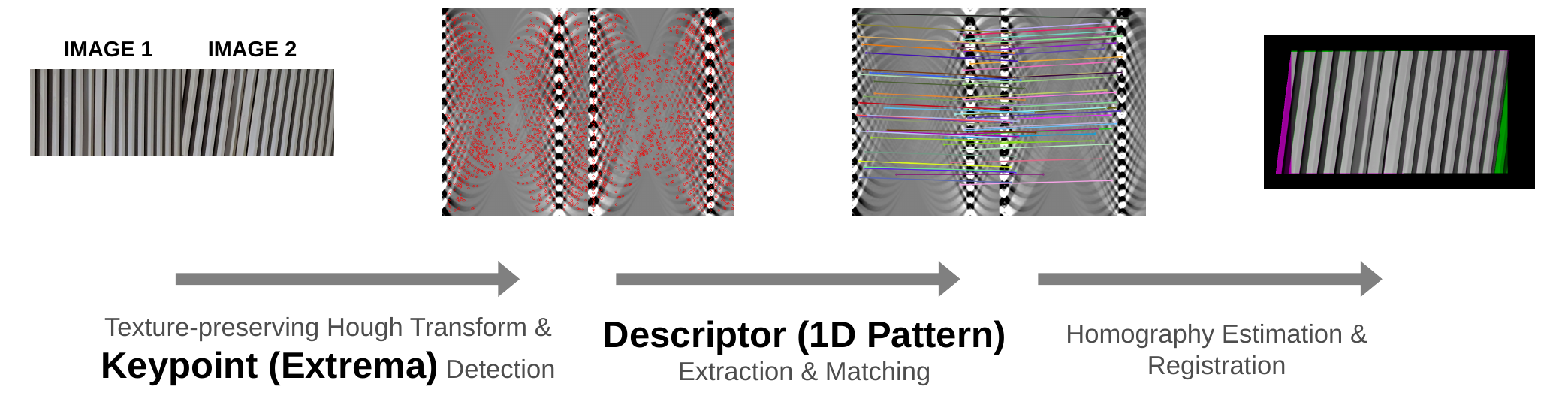}
      \captionof{figure}{Overview of the proposed image registration method, HOME. To localize line structures, HOME converts images to Hough space and conducts registration there. It simply uses Hough-space extrema as keypoints. Because the Hough transform consists of a summation along lines, it can suppress noise and improve the stability of extrema. Therefore, in Hough space, extrema are expected to be stable enough to serve as keypoints. Because rotational invariance is one of the most important properties for video frame registration, many image-space keypoint descriptors require elaborate designs to achieve it. In contrast, the HOME descriptor is a one-dimensional pattern along the radial direction, which is strictly invariant under rotation and translation of the original image; thus rotational and translational invariance is guaranteed automatically. This ultra-lightweight keypoint detector and descriptor enable HOME to achieve real-time performance while maintaining sufficient accuracy and robustness. In this example, image 1 (green) is warped to align with image 2 (purple) based on the estimated homography. The monochrome areas indicate successful registration.}
      \label{fig:overview}
  \end{center}
}
\makeatother

\DeclareMathOperator*{\argmin}{arg\,min}

\begin{document}
\pagestyle{plain}
\thispagestyle{plain}

\title{HOME: Robust Hough-space Matching Method for Structured and Textureless Videos}
\author{\IEEEauthorblockN{Masaki Satoh}
\IEEEauthorblockA{
\textit{Morpho, Inc.}\\
Tokyo, Japan \\
m-satoh@morphoinc.com}
}

\maketitle

\IEEEpeerreviewmaketitle

\begin{abstract}
   Visual front-ends for robotic localization typically rely on point-based features such as Oriented FAST and Rotated BRIEF (ORB), which frequently fail in structured environments dominated by strong linear structures or textureless surfaces. While line-based Simultaneous Localization and Mapping (SLAM) systems mitigate this by utilizing line segments, conventional line extraction and description algorithms are computationally prohibitive for real-time edge robotics. To address this fundamental bottleneck, we propose HOME (Hough-space One-dimensional Matching of Extrema), an ultra-lightweight, training-free feature matching framework. HOME transforms images into Hough space, mapping global linear structures to stable local extrema, which serve as keypoints, thereby reformulating complex line matching into highly efficient one-dimensional point matching. The proposed 1D radial descriptor mathematically guarantees rotational and translational invariance without the overhead of explicit orientation estimation. As a proof of concept to validate the matching accuracy and efficiency of HOME, this paper focuses on homography estimation. Extensive evaluations demonstrate that HOME achieves robust registration in challenging scenarios where point-based methods fail, operating at a much faster speed than existing line-based methods. Extending this robust matching engine to full 3D pose estimation remains a highly promising future direction.
\end{abstract}


\section{Introduction}

Visual front-ends play a critical role in robotics applications, such as Visual Odometry (VO), Simultaneous Localization and Mapping (SLAM), and self-positioning of autonomous mobile robots. The standard procedure for establishing frame-to-frame correspondences relies on point-based keypoint detection and description algorithms, such as Scale-Invariant Feature Transform (SIFT) \cite{lowe2004sift} and Oriented FAST and Rotated BRIEF (ORB) \cite{rublee2011orb}. In robotics, ORB is particularly favored for its real-time performance on embedded CPUs. However, point-based algorithms inherently focus on local patches and often fail in human-made environments dominated by linear structures (e.g., corridors, shutters, and pillars). In such environments, local point features become highly ambiguous, frequently leading to severe mismatches and catastrophic tracking failures in standard navigation pipelines.

To overcome the fragility of point-based methods, line-based SLAM and VO systems, such as PL-SLAM \cite{pumarola2017pl}, have been extensively studied. These systems extract structural line segments using algorithms such as the Line Segment Detector (LSD) \cite{von2010lsd} and match them using descriptors such as the Line Band Descriptor (LBD) \cite{zhang2013lbd}. While line segments provide significantly greater stability under illumination changes and lack of texture, the line extraction and description processes introduce severe computational overhead. The complex geometric verification required to handle line fragmentation makes these methods too heavy for resource-limited edge robots, precluding real-time performance.

To resolve this computational bottleneck, we reformulate the correspondence problem in Hough space and propose HOME (Hough-space One-dimensional Matching of Extrema), a novel ultra-lightweight matching framework natively designed for Hough space. HOME uses Hough-space extrema as keypoints and introduces a one-dimensional radial pattern as a descriptor. Because the Hough transform is summation-based, it naturally suppresses image noise and stabilizes the extrema. Furthermore, the proposed radial descriptor mathematically guarantees rotational and translational invariance without explicit orientation estimation. \figurename~\ref{fig:overview} illustrates the overall concept of HOME.

The primary contribution of HOME is an extremely fast and robust engine for global line correspondences. As a proof of concept to validate the accuracy and computational efficiency of this matching engine, this paper focuses on homography estimation. Integrating the proposed high-speed Hough matching into complete 3D state-estimation frameworks is an important future direction. Operating faster than real-time requirements, HOME offers a dependable, deterministic alternative to heavy line-based pipelines.

\section{Related Work}

\subsection{Robust Visual Front-ends in Robotics}
Visual front-ends for robotic localization rely heavily on establishing accurate correspondences across consecutive frames. Point-based methods, typified by ORB-SLAM \cite{mur2015orb}, use ORB features due to their speed. While highly successful in feature-rich environments, these methods are notoriously fragile in structured, man-made environments because they cannot detect a sufficient number of reliable keypoints on linear structures, leading to feature-matching failures.

To mitigate this, line-based SLAM and VO systems, such as PL-SLAM \cite{pumarola2017pl}, leverage structural line segments detected by algorithms like LSD \cite{von2010lsd}. Line segments offer greater stability under illumination changes and in textureless regions. However, describing and matching line segments using methods such as LBD \cite{zhang2013lbd} introduces significant computational overhead. The complex geometric verification required to handle line fragmentation often violates the real-time constraints of resource-limited robots. HOME bypasses this bottleneck by mapping the entire structural layout into Hough space, allowing structural line matching to be reformulated as a highly efficient 1D point matching problem.

\subsection{Hough-space Matching}
The Hough transform \cite{duda1972hough} is a classical method for detecting geometric primitives. The concept of performing feature matching directly in a transformed domain was explored in Hough-SIFT \cite{satoh2026hough}, which demonstrated that keypoints can be extracted from Hough images to handle strong linear structures. Although robust, Hough-SIFT is computationally expensive because it applies the full SIFT pipeline to Hough images. In contrast, HOME leverages the intrinsic geometric properties of the Hough transform to design a native 1D descriptor, drastically reducing execution time while maintaining the robustness inherent to Hough-domain processing.

\section{Proposed Method}

This section describes HOME in detail. We first review the texture-preserving Hough transform, then present keypoint detection and descriptor design, and finally explain homography estimation.

\subsection{Texture-Preserving Hough Transform}

Let $(x, y)$ denote the coordinates in the original image $I(x, y)$, and parameterize the Hough space by $\theta$ and $r$, where $\theta$ is the angle of the normal vector $\bm{n}(\theta) = (\cos\theta, \sin\theta)$ of a line and $r$ is the distance along $\bm{n}(\theta)$ from the origin to the line. The line equation can be written as $x\cos\theta + y\sin\theta - r = 0$. The texture-preserving Hough transform is defined as follows:
\begin{align}
   J(\theta,r) =
   \left[
   \sum_{(x,y) \in \bm{l}(\theta,r)}
   \nabla I(x,y)
   \right] \cdot \bm{n}(\theta).
   \label{eq:hough_transform}
\end{align}
To improve stability, we apply Gaussian smoothing to $J(\theta,r)$. For line summation, we sum pixels near each line instead of interpolating exact on-line values. The $r$-derivative of the Radon transform \cite{radon1917} is mathematically equivalent to this definition, and it may seem more efficient to first compute the Radon transform and then take the $r$-derivative. We use \eqref{eq:hough_transform} because it is more robust to quantization errors introduced by the nearest-neighbor approximation. Because local image intensity is approximately planar, quantization has a smaller effect on gradients.

The purpose of this transform is to enhance linear structures in the image while preserving texture information. Although many variations are possible, this definition has three important properties. First, it consists of simple summation and can therefore suppress noise and improve registration stability. This property makes Hough-based methods robust not only to linear structures but also to noise. Second, the Hough transform can be computed by summing gradients along lines and then taking the inner product with the normal vector, which is computationally efficient. With SIMD (e.g., NEON and AVX) and multithreading, real-time processing is feasible, and GPU acceleration can further improve throughput. Third, we can use the sign of $J(\theta,r)$ to reduce the number of matching combinations. Along the normal vector, the sign of $J(\theta,r)$ reflects whether a line structure is bright-to-dark or dark-to-bright in the image, and this property can be treated as invariant under wide varieties of image transformations, including translations and rotations. Matching only keypoints with the same sign of $J(\theta,r)$ reduces the number of combinations.

\subsection{Keypoint Detection}
\label{sec:keypoint_detection}

Image-space extrema are often non-distinctive and noise-sensitive. In Hough space, this changes: the line-summation process in \eqref{eq:hough_transform} suppresses noise and stabilizes extrema. In addition, geometrically, Hough extrema correspond to line structures, making them strong keypoint candidates. HOME therefore uses Hough-space extrema as keypoints. Extrema extraction is lightweight compared with SIFT and conventional corner/blob detectors, reducing computational cost. Another advantage is sub-pixel localization. Let $(\theta_0,r_0)$ be an integer extremum in Hough space. By applying linear filters to $J(\theta,r)$, we estimate first and second derivatives, and the sub-pixel location $(\hat{\theta}_0,\hat{r}_0)$ is obtained as:
\begin{align}
   \begin{bmatrix}
      \hat{\theta}_0 \\ \hat{r}_0
   \end{bmatrix}
   =
   \begin{bmatrix}
      \theta_0 \\ r_0
   \end{bmatrix}
   - H^{-1}(\theta_0,r_0)
   \begin{bmatrix}
      J_\theta(\theta_0,r_0) \\
      J_r(\theta_0,r_0)
   \end{bmatrix},
\end{align}
where $H(\theta_0,r_0)$ is the Hessian matrix of $J(\theta,r)$ at $(\theta_0,r_0)$, and $J_\theta$ and $J_r$ are the first derivatives of $J(\theta,r)$ along the $\theta$ and $r$ directions, respectively.

To manage computational cost, we select keypoints using a saliency measure rather than using all extrema. We rank keypoints by saliency and select a fixed number for matching and homography estimation. We use $|\mathrm{det}(H(\theta_0,r_0))/\mathrm{tr}(H(\theta_0,r_0))|$ for this ranking. This value can be estimated as
\begin{align*}
   \frac{\mathrm{det}(H)}{\mathrm{tr}(H)}
   =
   \frac{1}{1/\lambda_\mathrm{H} + 1/\lambda_\mathrm{L}}
   \simeq \lambda_\mathrm{L},
\end{align*}
where $\lambda_\mathrm{H}$ and $\lambda_\mathrm{L}$ are the high and low eigenvalues of $H$, respectively, and we assume $|\lambda_\mathrm{H}| \gg |\lambda_\mathrm{L}|$. For a prominent extremum, $\lambda_\mathrm{L}$ (the smaller curvature) should be large. Therefore, $|\mathrm{det}(H)/\mathrm{tr}(H)|$ is a useful saliency measure.

We analyze the robustness of Hough-space extrema for $\theta=0$; this is without loss of generality because the coordinate system can be rotated accordingly. Equation \eqref{eq:hough_transform} and its $r$ derivative become:
\begin{align}
   J(0,r) = \sum_y I_{x}(r,y), \quad
   J_r(0,r) = \sum_y I_{xx}(r,y).
   \label{eq:hough_transform_theta0}
\end{align}
To obtain the $\theta$ derivative, we rotate the image $I(x,y)$ in the reverse direction instead of rotating the summation line $\bm{l}$ and normal vector $\bm{n}$. Straightforward calculation gives
\begin{align}
   J_\theta(0,r) = - \sum_y I_{xx}(r,y) y + r \sum_y I_{xy}(r,y).
\end{align}
For $J(\theta,r)$ to be extremal at $(0,r_0)$, the conditions $J_\theta(0,r_0) = 0$ and $J_r(0,r_0) = 0$ must be satisfied. These are necessary, but not sufficient, conditions; we do not investigate this issue further in this paper. To study extrema properties, we consider a separable local image model near the line, $I(x,y) = f(x) g(y)$. Then the conditions reduce to $f''(r)=0$ and $\sum_y g'(y)=0$.

Next, we estimate $J(0,r_0)$. Since $f''(r)=0$, $f(x)$ is locally linear: $f(x)=a(x-r)+b$. Assuming $\sum_y g(y)=L$ (line length $L$), \eqref{eq:hough_transform_theta0} gives $J(0,r)=aL$. We then add independent and identically distributed (i.i.d.) Gaussian noise $n(x,y)\sim\mathcal{N}(0,\sigma^2)$: $I(x,y)=f(x)g(y)+n(x,y)$, with $\langle n(x,y)\rangle=0$ and $\langle n(x,y)n(x',y')\rangle=\sigma^2\delta_{x,x'}\delta_{y,y'}$. Decomposing $J(0,r_0)=J_\mathrm{S}(0,r_0)+J_\mathrm{N}(0,r_0)$, where $J_\mathrm{S}(0,r_0)=aL$ and $J_\mathrm{N}(0,r_0)=\sum_y n_x(r_0,y)$, the signal-to-noise ratio (SNR) is
\begin{align}
   \mathrm{SNR} = \frac{J_\mathrm{S}(0,r_0)^2}{\langle J_\mathrm{N}(0,r_0)^2 \rangle} = 2L \frac{a^2}{\sigma^2},
\end{align}
where $n_x(r_0,y)$ is evaluated using a central-difference numerical derivative: $n_x(r_0,y) = (n(r_0+1,y) - n(r_0-1,y))/2$.

For example, with $L=640$, $a=1$ (weak line structure), and $\sigma=10$ (relatively large noise), the estimated SNR is $12.8$, which is still high for stable keypoint detection. This supports the view that summation in \eqref{eq:hough_transform} improves extremum stability under noise.

\subsection{Keypoint Description}

Many keypoint descriptors have addressed how to achieve geometric invariance, notably translation, rotation, and scale invariance in image space. Translation invariance is automatically achieved by using local patterns around keypoints, and scale invariance is less critical in video processing because scale changes are usually small. However, rotational invariance is crucial, and each descriptor algorithm has its own design to achieve it. For example, SIFT achieves it by rotating the local pattern according to the dominant gradient orientation around keypoints, and ORB achieves it by rotating the local pattern according to the intensity centroid around keypoints.

As in keypoint detection, the situation is very different in Hough space. We investigate how Hough-space patterns behave under rotational and translational transformations of the original image. Let homogeneous image coordinates $(x, y, 1)^\mathrm{T}$ be transformed to $(x', y', 1)^\mathrm{T}$ by a homography $H$: $(x', y', 1)^\mathrm{T} \sim H (x, y, 1)^\mathrm{T}$, where homography $H$ consists of rotation $\alpha$ followed by translation $(t_x, t_y)$. Because the homogeneous representation of a line is transformed by the inverse transpose of the point homography matrix, the line $(\cos\theta,\sin\theta,-r)^\mathrm{T}$ is transformed to $(\cos\theta',\sin\theta',-r')^\mathrm{T} \sim H^{-\mathrm{T}} (\cos\theta,\sin\theta,-r)^\mathrm{T}$. A straightforward calculation gives that the Hough-space point $(\theta, r)$ is transformed to $(\theta + \alpha, r + t_x \cos(\theta + \alpha) + t_y \sin(\theta + \alpha))$. This mapping is neither a simple translation/rotation nor a homography, which explains why SIFT is not an optimal descriptor there.

Consider two lines with the same normal vector: $(\theta_0,r_1)$ and $(\theta_0,r_2)$. Image rotation by $\alpha$ and translation by $(t_x, t_y)$ transform them as $(\theta_0,r_i) \rightarrow (\theta'_0,r'_i) \equiv (\theta_0 + \alpha, r_i + t_x \cos(\theta + \alpha) + t_y \sin(\theta + \alpha))$, where $i \in \{1, 2\}$. Hence, two parallel lines in the original image are transformed to another pair of parallel lines, and the distance between them is preserved under the transformation, or $r'_2 - r'_1 = r_2 - r_1$. This can also be understood by geometric considerations. Consider two parallel lines in the original image. It is clear that, by rotation and translation of the image, they are transformed to another pair of parallel lines, and the distance between them is preserved. In other words, Hough-space patterns along the radial direction are invariant under rotation and translation of the original image. Therefore, by using these patterns as descriptor vectors, we can automatically and strictly guarantee rotational and translational invariance. This descriptor definition is extremely simple and lightweight. With precomputed shrunk and (if needed) transposed images, we only need to copy one-dimensional patterns around keypoints. It is much faster than the SIFT descriptor and even faster than ORB, which is designed for fast computation.

For descriptor construction, we quantize the Hough image to 8-bit unsigned integers, crop a $2 \times 32$ patch around each keypoint, and apply a $1/2$ shrink to obtain a $1 \times 16$ HOME descriptor. Descriptor distance is the $L_1$ norm, and matching is brute-force with Lowe's ratio test \cite{lowe2004sift}.

\subsection{Homography Estimation}

After keypoint detection and matching, we estimate the homography between two images. Let $H$ map coordinates in the first image to those in the second image. We estimate $H$ from matched Hough-space keypoint pairs using the inverse-transpose line transform.

For the cost function, we use the difference of the homogeneous line coordinates $(a,b,c)^\mathrm{T}$ with the normalization $a^2+b^2=1$. This definition approximates the difference in the Hough space \cite{satoh2026hough}. Let $(\theta^i_1, r^i_1)$ and $(\theta^i_2, r^i_2)$ be the coordinates of the $i$-th matched keypoints in the Hough space. Then, we define the transfer error $\bm{\epsilon}^i(H)$ of the $i$-th matched pair as:
\begin{align}
   \bm{\epsilon}^i(H) \equiv
   \frac{1}{\sqrt{(a^i_1)^2 + (b^i_1)^2}}
   \begin{bmatrix}
      a^i_1 \\ b^i_1 \\ c^i_1
   \end{bmatrix}
   -
   \begin{bmatrix}
      \cos\theta^i_2 \\ \sin\theta^i_2 \\ -r^i_2
   \end{bmatrix},
   \label{eq:transfer_error}
\end{align}
where $(a^i_1, b^i_1, c^i_1)^\mathrm{T} \equiv H^{-\mathrm{T}} (\cos\theta^i_1, \sin\theta^i_1, -r^i_1)^\mathrm{T}$. We estimate $H$ by solving the following optimization problem:
\begin{align}
  \hat{H} = \argmin_H \sum_i \bm{\epsilon}^i(H)^\mathrm{T} \bm{\epsilon}^i(H).
\end{align}

For solving this kind of optimization problem with outliers, a standard approach is RANSAC \cite{fischler1981ransac} followed by Levenberg-Marquardt refinement. For real-time video, however, RANSAC has unpredictable runtime and unnecessary robustness when consecutive-frame homographies are usually close to identity. We therefore use a deterministic scheme: initialize $H$ as identity, iteratively halve the inlier threshold, and re-estimate $H$ from inliers by Levenberg-Marquardt.

\section{Experiments and Discussion}

In this section, we evaluate the performance of HOME by comparing it with ORB, Accelerated-KAZE (AKAZE) \cite{alcantarilla2013}, and LSD/LBD. ORB is a fast registration algorithm designed for real-time performance on embedded systems. AKAZE is another fast algorithm that prioritizes accuracy. These two are point-based algorithms designed for texture-rich situations. We check if HOME can achieve comparable performance in such situations. In contrast, LSD/LBD is a line-based registration algorithm that is robust to linear structures and is therefore a natural benchmark for HOME.

For ORB, AKAZE, LSD/LBD, and their homography estimation, we use the OpenCV implementations. We use default settings for ORB and AKAZE with Lowe's ratio test \cite{lowe2004sift} (threshold 0.75), except ORB's feature count is set to 2000. For LSD/LBD, we use the default settings and the same $0.75$ ratio test, followed by a geometric consistency check that discards matches if the relative length difference between the two corresponding line segments exceeds 50\%. The homography estimation is performed using the OpenCV function \texttt{findHomography}, which implements RANSAC and Levenberg-Marquardt refinement, with the default settings. For LSD/LBD homography estimation, we extract three point correspondences per line segment: the two endpoints and the midpoint.

In HOME, we convert input images to Hough images with resolution $(480+120)\times640$, where $480$ corresponds to the $\theta$ axis and $640$ to the $r$ axis, and $+120$ is padding on the $\theta$ axis to avoid boundary effects. We detect 1000 positive and 1000 negative keypoints in HOME (matching ORB's 2000 total), where positive and negative denote the sign of the Hough response. The standard deviation of the Gaussian smoothing applied before keypoint detection is set to $1.3$ pixels. The final threshold for deterministic inlier selection is set to $1.0$ pixels.

\subsection{Video Sequences}

\begin{figure}[!htb]
   \centering
   \begin{minipage}{0.2\linewidth}
      \centering
      \includegraphics[width=\linewidth]{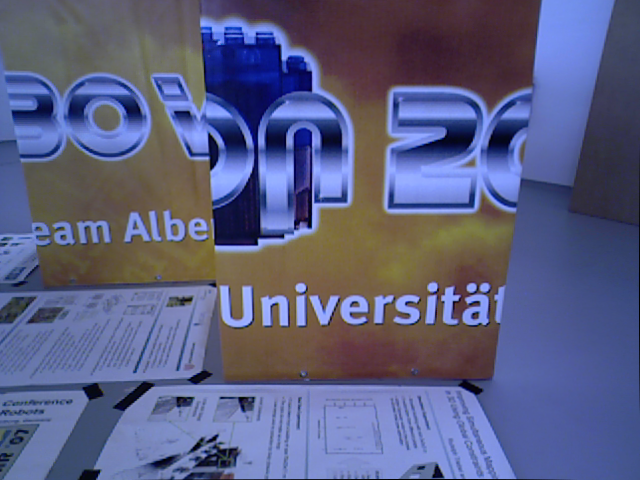}
      \\ (a)
   \end{minipage}
   \begin{minipage}{0.2\linewidth}
      \centering
      \includegraphics[width=\linewidth]{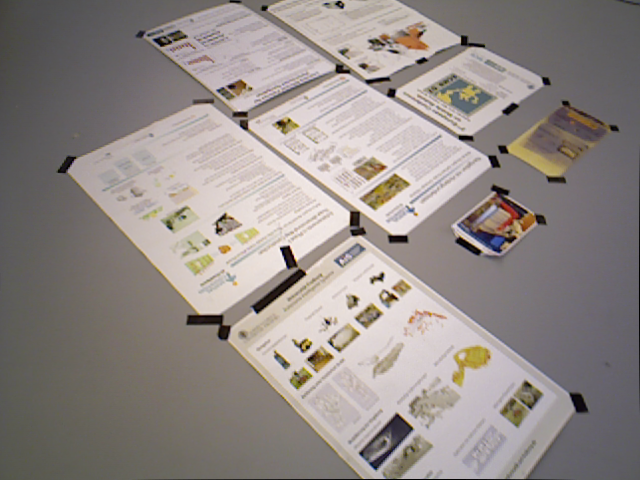}
      \\ (b)
   \end{minipage}
   \begin{minipage}{0.2\linewidth}
      \centering
      \includegraphics[width=\linewidth]{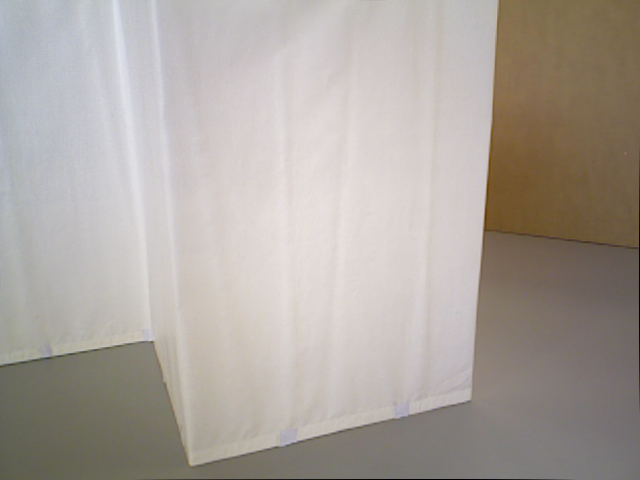}
      \\ (c)
   \end{minipage}
   \begin{minipage}{0.2\linewidth}
      \centering
      \includegraphics[width=\linewidth]{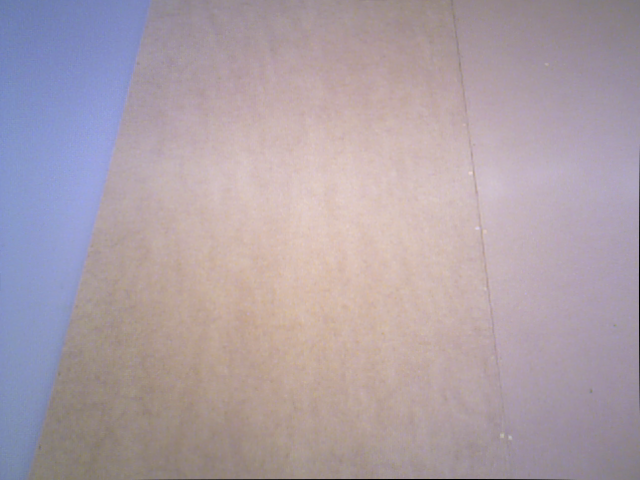}
      \\ (d)
   \end{minipage}
   \caption{Sample frames from the ``Structure vs. Texture'' category of the TUM dataset. (a) Structure/Texture. (b) No structure/Texture. (c) Structure/No texture. (d) No structure/No texture.}
   \label{fig:sample_frames_tum}
\end{figure}

HOME is a registration algorithm primarily designed for video sequences with linear structures. Therefore, we evaluate it on the ``Structure vs. Texture'' category of the TUM dataset \cite{alcantarilla2013}, which contains ``Structure/Texture,'' ``No structure/Texture,'' ``Structure/No texture,'' and ``No structure/No texture'' sequences. We show sample frames from these sequences in \figurename~\ref{fig:sample_frames_tum}. The ``Structure/Texture'' and ``No structure/Texture'' sequences verify that HOME maintains performance where standard methods excel. The ``Structure/No texture'' sequences contain strong linear structures with weak textures, challenging point-based methods. The ``No structure/No texture'' sequences are difficult for all methods.

Since ground-truth homography is unavailable for this dataset, we evaluate registration accuracy using photometric error between consecutive frames rather than geometric error. Photometric error is the root-mean-square error (RMSE) between the warped previous frame and current frame in the overlapping region. Frames with failed homography estimation are excluded to avoid skewing the distribution. We define homography-estimation failure as a case where the number of inliers for final homography estimation is less than 20, or the determinant of the normalized homography, $H/H(2,2)$, is less than or equal to zero, which indicates an unrealistic transform, such as coordinate flipping.

We summarize registration performance in \tablename~\ref{tab:results_tum}. We also report failure rate, defined as the percentage of frames where homography estimation fails, together with the number of detected keypoints (num. keypoints), matched keypoints (num. matches), and inliers used for final homography estimation (num. inliers). Since space is limited, we report only the average of each metric across frames and videos. Note that keypoints indicate line segments for LSD/LBD.

\begin{table*}[!htb]
   \centering
   \caption{Summary of registration performance of TUM's structure vs. texture dataset.}
   \label{tab:results_tum}
   \begin{tabular}{l|rrrr|rrrr|rrrr}
      & \multicolumn{4}{c|}{(Structure\textbar No structure)/Texture} & \multicolumn{4}{c|}{Structure/No texture} &
      \multicolumn{4}{c}{No structure/No texture} \\
      & \multicolumn{4}{c|}{(4 Sequences/4180 Pairs)} & \multicolumn{4}{c|}{(2 Sequences/1894 Pairs)} &
      \multicolumn{4}{c}{(2 Sequences/1597 Pairs)} \\
      & ORB & AKAZE & LBD & HOME & ORB & AKAZE & LBD & HOME & ORB & AKAZE & LBD & HOME \\
      \hline
      \hline
      RMSE & $7.699$ & $7.375$ & $7.714$ & $7.241$ & $8.351$ & $5.574$ & $4.438$ & $4.024$
      & $20.345$ & --- & $9.854$ & $5.479$ \\
      Failure rate & $1.9\%$ & $1.8\%$ & $0.3\%$ & $0\%$ & $76.3\%$ & $87.7\%$ & $42.5\%$ & $0\%$
      & $99.9\%$ & $100\%$ & $99.3\%$ & $0.3\%$ \\
      \hline
      Num. keypoints & $789.5$ & $461.0$ & $399.1$ & $1998.7$ & $38.5$ & $14.7$ & $21.5$ & $2000$
      & $4.5$ & $0.4$ & $8.4$ & $2000$ \\
      Num. matches & $431.8$ & $366.5$ & $243.1$ & $1355.4$ & $17.2$ & $12.5$ & $12.2$ & $765.5$
      & $2.1$ & $0.1$ & $2.9$ & $541.2$ \\
      Num. inliers & $387.5$ & $352.6$ & $204.2$ & $1282.4$ & $14.3$ & $11.6$ & $8.4$ & $572.6$
      & $1.1$ & $0.0$ & $1.1$ & $434.4$ \\
   \end{tabular}
\end{table*}

For the ``Structure/Texture'' and ``No Structure/Texture'' sequences, all methods perform well. HOME achieves the best RMSE and failure rate. For the ``Structure/No texture'' sequences, only HOME achieves complete registration. LSD/LBD often fails because the number of prominent line segments is too small to estimate a homography. Notably, HOME can register images even in the ``No structure/No texture'' sequences.

In this experiment, HOME detects the maximum 2000 keypoints for nearly all frames. These detections are genuine rather than noise artifacts, as evidenced by the high number of inliers. This trend reflects the noise robustness of Hough-space extrema discussed in Section \ref{sec:keypoint_detection}. This may explain why HOME outperforms in textured sequences despite being designed for linear structures; further investigation is reserved for future work.

\begin{figure}[!htb]
   \centering
   \begin{minipage}{0.23\linewidth}
      \centering
      \includegraphics[width=\linewidth]{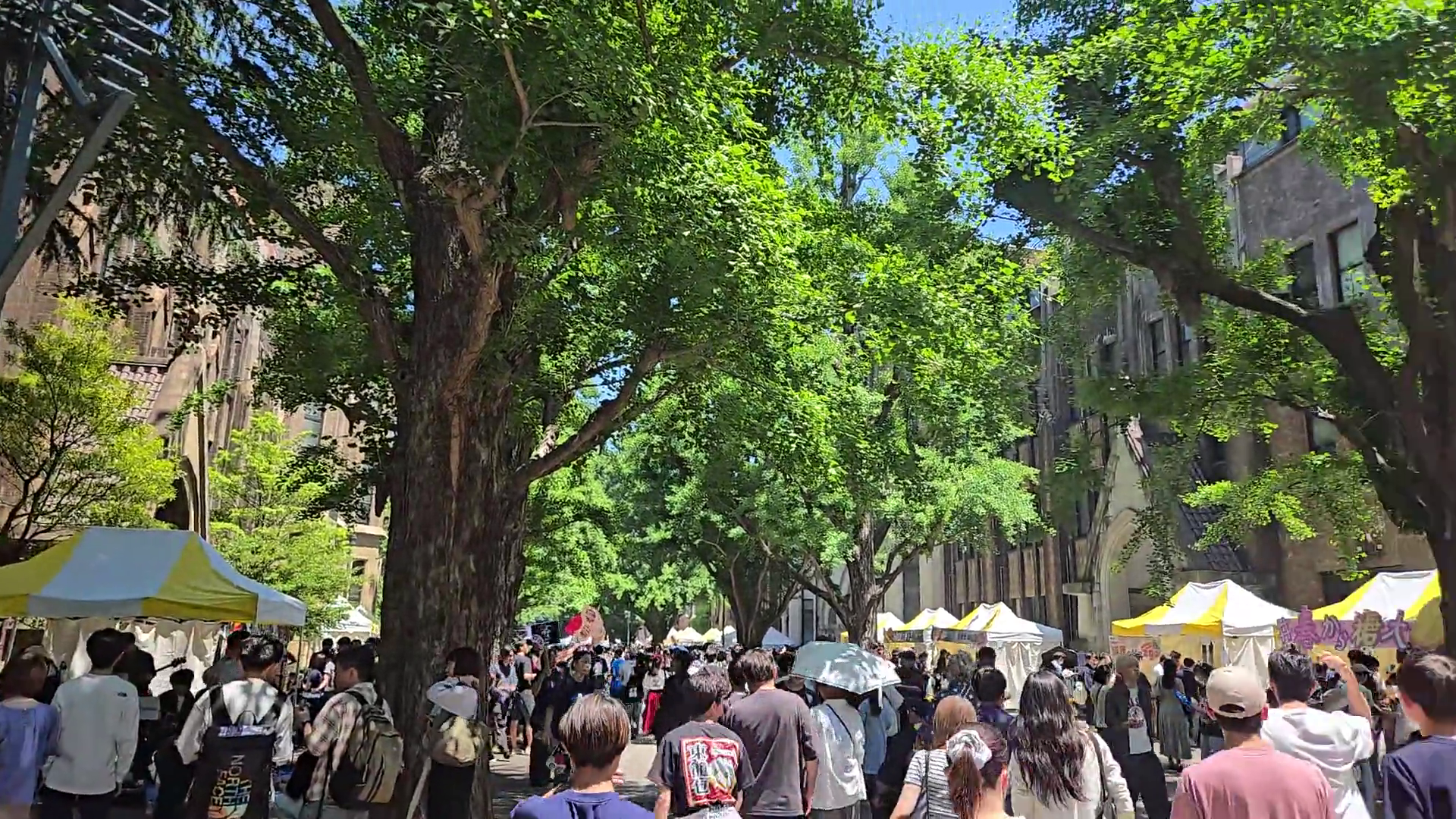}
   \end{minipage}
   \begin{minipage}{0.23\linewidth}
      \centering
      \includegraphics[width=\linewidth]{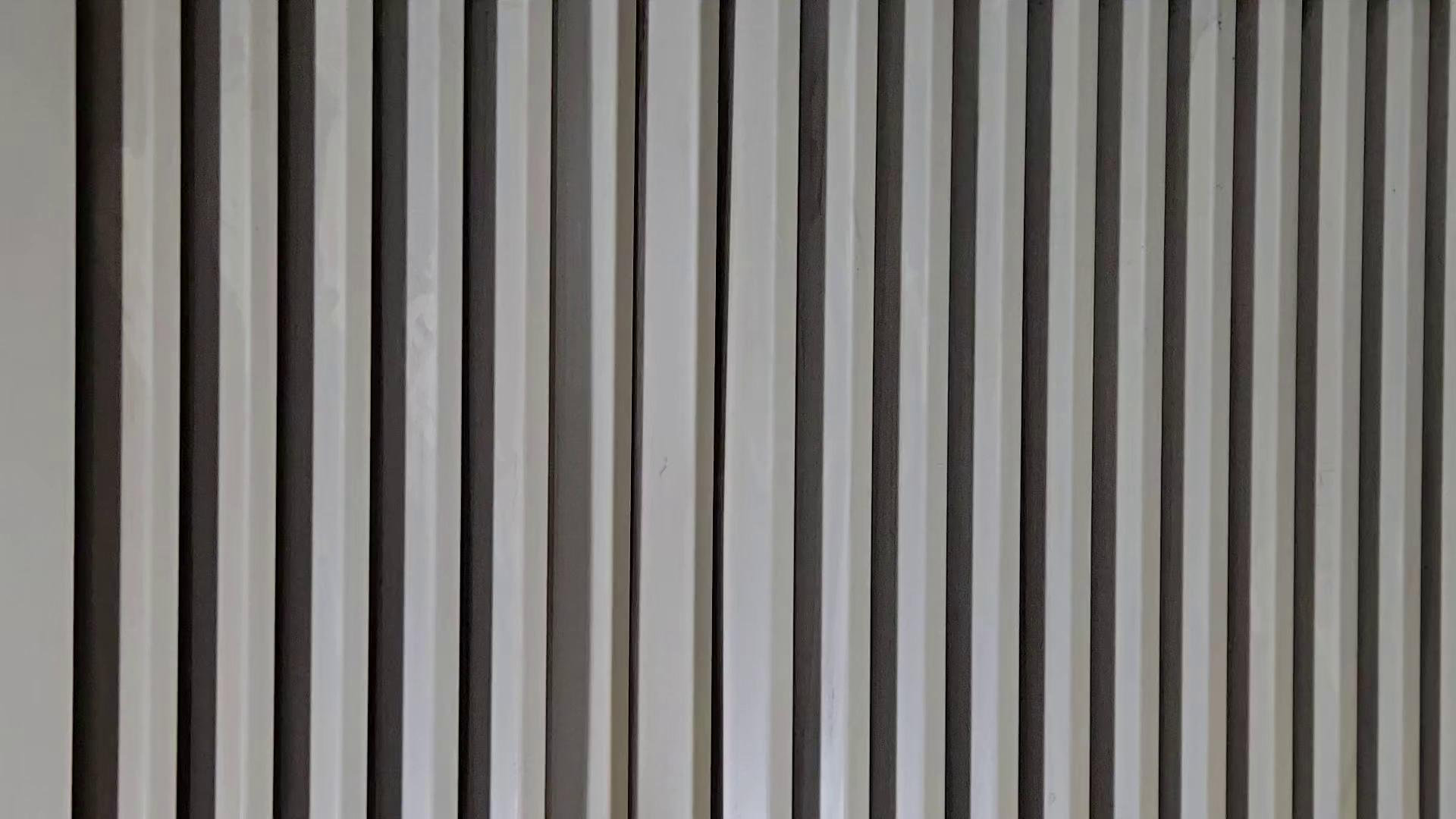}
   \end{minipage}
   \begin{minipage}{0.23\linewidth}
      \centering
      \includegraphics[width=\linewidth]{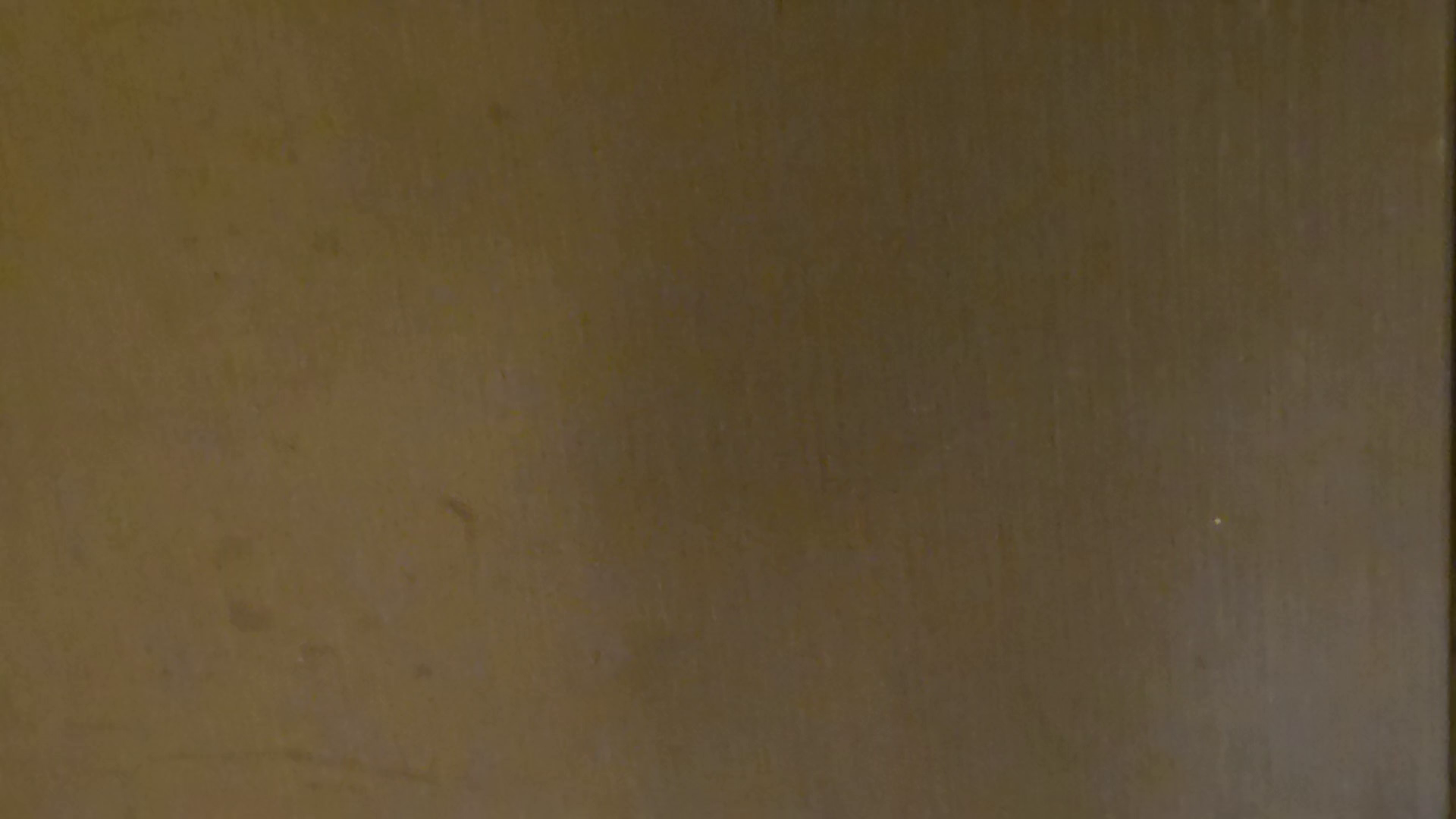}
   \end{minipage}
   \caption{Sample frames from additional video clips. (Left) Normal video. (Center) Linear video. (Right) Textureless video.}
   \label{fig:sample_frames_ours}
\end{figure}

Because the number of sequences is limited in the TUM dataset, we prepared some additional video clips of three categories to support our evaluation. We show sample frames in \figurename~\ref{fig:sample_frames_ours}. The first category is ``normal'' videos with rich textures for all methods. The second category is ``linear'' videos where line-based methods should excel. The third category is challenging ``textureless'' videos. The resolution of all videos is $640\times360$.

We summarize registration performance in \tablename~\ref{tab:results_ours}, where we report the same metrics as in \tablename~\ref{tab:results_tum}. Results align with \tablename~\ref{tab:results_tum}. The main difference is that LSD/LBD performs better on our linear videos than the TUM "Structure/No texture" category, owing to more prominent line segments in our dataset.

\begin{table*}[!htb]
   \centering
   \caption{Summary of registration performance.}
   \label{tab:results_ours}
   \begin{tabular}{l|rrrr|rrrr|rrrr}
      & \multicolumn{4}{c|}{Normal} & \multicolumn{4}{c|}{Linear} & \multicolumn{4}{c}{Textureless} \\
      & \multicolumn{4}{c|}{(5 Sequences/595 Pairs)} & \multicolumn{4}{c|}{(8 Sequences/952 Pairs)} & \multicolumn{4}{c}{(3 Sequences/357 Pairs)} \\
      & ORB & AKAZE & LBD & HOME & ORB & AKAZE & LBD & HOME & ORB & AKAZE & LBD & HOME \\
      \hline
      \hline
      RMSE & $16.878$ & $16.334$ & $18.134$ & $16.203$ & $6.683$ & $3.686$ & $5.056$ & $3.481$
      & $7.137$ & --- & --- & $2.364$ \\
      Failure rate & $0\%$ & $0\%$ & $0\%$ & $0\%$ & $73.3\%$ & $62.6\%$ & $16.1\%$ & $0\%$
      & $97.8\%$ & $100\%$ & $100\%$ & $0.8\%$ \\
      \hline
      Num. keypoints & $2000$ & $1263.3$ & $405.5$ & $2000$ & $83.2$ & $55.8$ & $91.5$ & $1998.6$
      & $19.3$ & $0.0$ & $1.7$ & $2000.0$ \\
      Num. matches & $1286.5$ & $1012.3$ & $240.4$ & $1251.5$ & $43.4$ & $33.9$ & $42.7$ & $612.7$
      & $1.2$ & $0.0$ & $0.4$ & $680.0$ \\
      Num. inliers & $1189.1$ & $962.1$ & $196.3$ & $1206.6$ & $33.1$ & $27.9$ & $31.4$ & $425.8$
      & $1.2$ & $0.0$ & $0.1$ & $608.3$ \\
   \end{tabular}
\end{table*}

Here, we measure runtime on the normal videos, where more keypoints are detected and the computational load is higher. Experiments were run on a laptop with an AMD Ryzen 7 7735U CPU and 32 GB RAM. We report average times for Hough transform (only for HOME), keypoint detection/description (KP detection/description), keypoint matching (KP matching), homography estimation, and total runtime (\tablename~\ref{tab:results_processing_time}). As expected, ORB is the fastest, but HOME is also real-time capable. Most HOME runtime is consumed by the Hough transform, which is suitable for GPU acceleration, while HOME's detection/description and matching are faster than ORB's. Since OpenCV's ORB is heavily optimized while HOME is not, significant speedup is possible. LSD/LBD demonstrates better robustness to linear structures than ORB and AKAZE, but its processing time is prohibitive for real-time applications.

\begin{table}[!htb]
   \centering
   \caption{Average processing time (ms).}
   \label{tab:results_processing_time}
   \begin{tabular}{lrrrr}
      & ORB & AKAZE & LBD & HOME \\
      \hline
      \hline
      Hough transform & --- & --- & --- & $7.741$ \\
      KP detection/description & $3.881$ & $20.468$ & $31.893$ & $3.272$ \\
      KP matching & $5.536$ & $2.852$ & $1.203$ & $3.434$ \\
      Homography estimation & $0.908$ & $0.766$ & $0.707$ & $1.896$ \\
      \hline
      Total & $10.486$ & $24.151$ & $34.099$ & $16.511$ \\
   \end{tabular}
\end{table}

Thus far, HOME outperforms LSD/LBD in every evaluated aspect, including RMSE, failure rate, the number of detected lines, and runtime. Therefore, it is a promising candidate to replace LSD/LBD in robotics applications such as SLAM. Moreover, it shows no degradation in texture-rich images compared with ORB and AKAZE, while remaining real-time capable.

\subsection{Synthetic Image Pairs}

Low photometric errors alone cannot verify whether the estimated homographies are geometrically correct. RMSE is a good alternative metric for texture-rich images, but it is unsuitable for linear or textureless images. Therefore, we conduct additional experiments using synthetic image pairs, where the ground-truth homography is available.

We prepare 20 images with strong linear structures at 854×480 resolution, designated the ``linear" dataset. Then, we warp each image using the ground-truth homography such that each image corner is randomly shifted within $p\in\{10,20,30,40,50,60\}$ pixels, where $p$ is the perturbation parameter defining the registration difficulty. By cropping the central $640\times360$ area from the original and warped images, followed by fixed Gaussian noise of $\sigma=2.5$ to both images, we obtain an image pair and the corresponding ground-truth homography for registration testing. We apply the noise because static noise patterns in the original image may be used as cues for registration, and we want to exclude this unrealistic behavior. We generate five pairs for each image, yielding 100 pairs in total for each dataset.

\begin{figure}[!htb]
   \centering
   \begin{minipage}{0.45\linewidth}
      \centering
      \includegraphics[width=\linewidth]{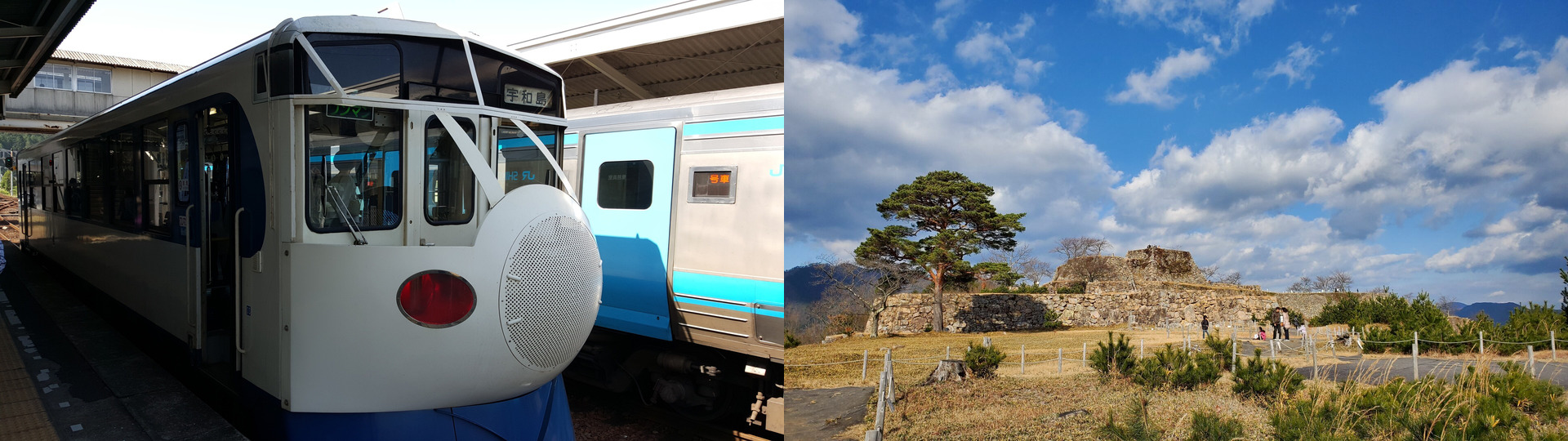}
   \end{minipage}
   \begin{minipage}{0.45\linewidth}
      \centering
      \includegraphics[width=\linewidth]{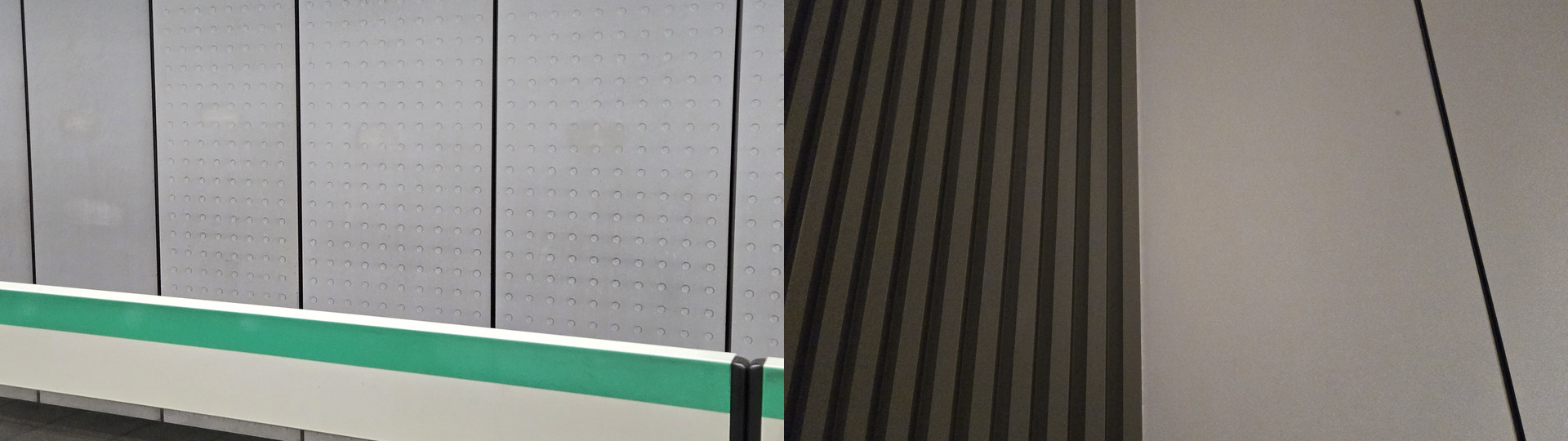}
   \end{minipage}
   \caption{Examples of images for generating synthetic image pairs. (Left) Normal dataset. (Right) Linear dataset.}
   \label{fig:dataset}
\end{figure}

We generate pairs for weak-texture robustness using the ``normal" dataset of 20 texture-rich images, applying the same procedure. We simulate weak-texture images by applying a texture-decay operation to the original and warped images. The texture-decay operation is defined as follows: $I'(x,y) = I(x,y) + \alpha( 128 - I(x,y) )$, where $I(x,y)$ and $I'(x,y)$ are the original and decayed images, respectively, and $\alpha$ is the texture-decay ratio. The larger $\alpha$ is, the weaker the texture becomes. We evaluate registration under texture decay $\alpha \in \{0, 0.4, 0.6, 0.8, 0.9, 0.95\}$ with fixed Gaussian noise ($\sigma=2.5$) and perturbation level ($p=20$). We show example images from the normal and linear datasets in \figurename~\ref{fig:dataset}.

\begin{figure}[!htb]
   \centering
   \begin{minipage}{0.45\linewidth}
      \centering
      \includegraphics[width=\linewidth]{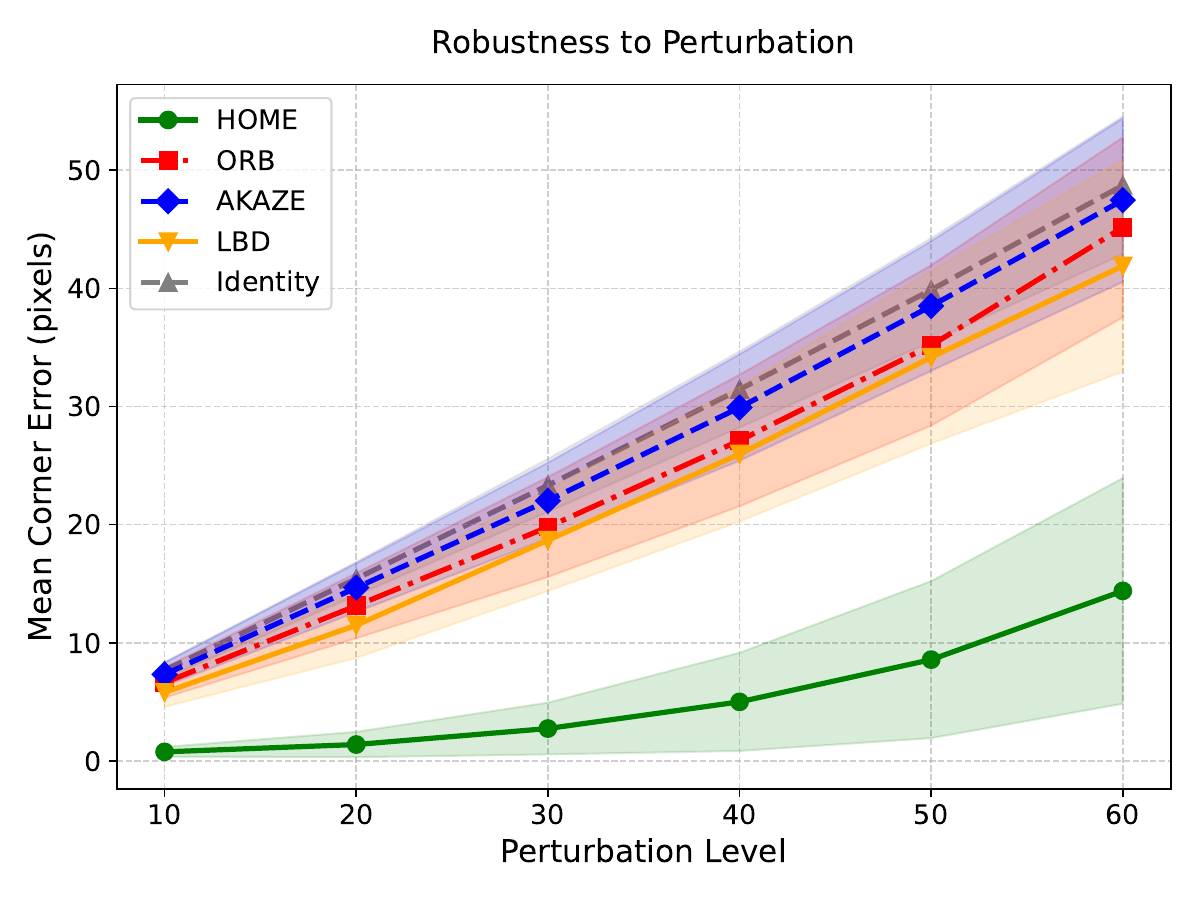}
   \end{minipage}
   \begin{minipage}{0.45\linewidth}
      \centering
      \includegraphics[width=\linewidth]{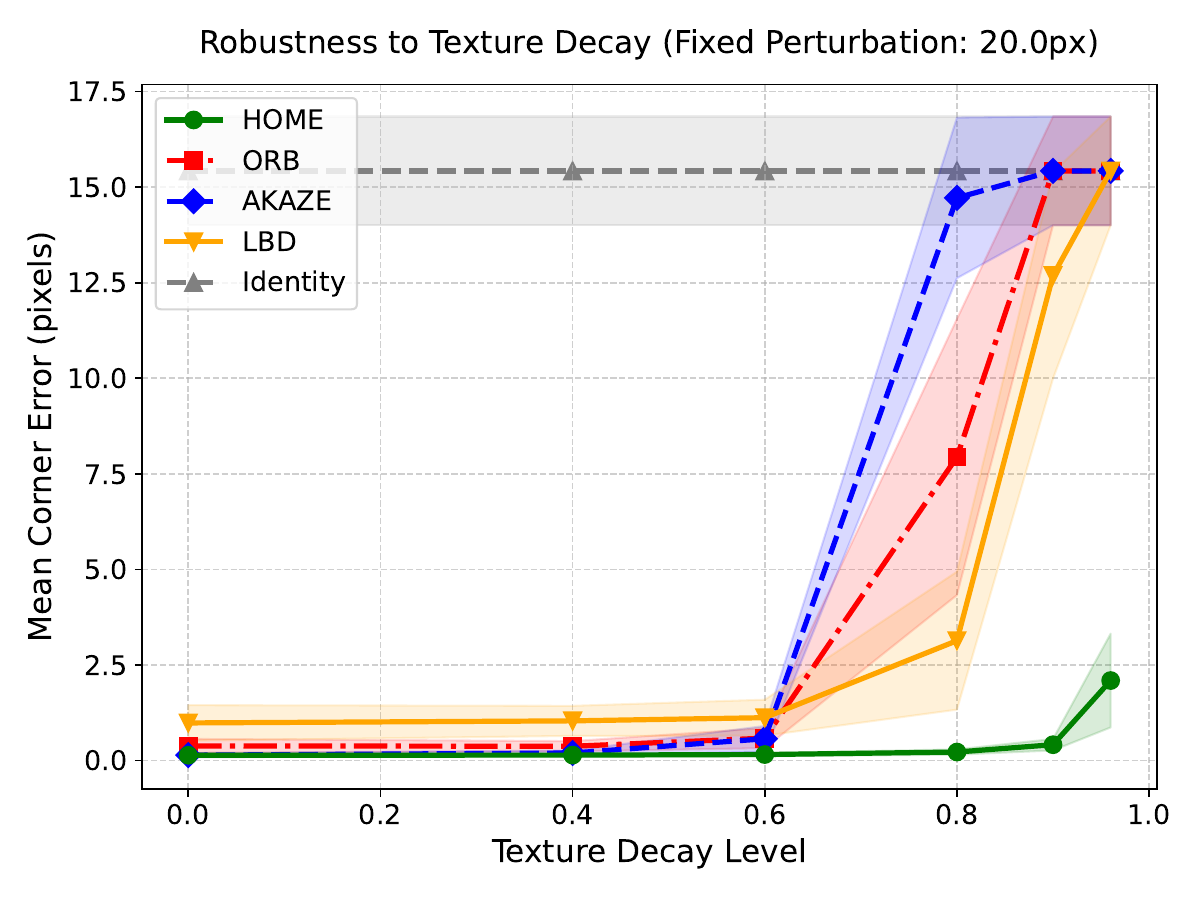}
   \end{minipage}
   \caption{Geometric error analysis of ORB, AKAZE, LSD/LBD, and HOME. (left) Geometric error under different perturbation levels with linear dataset. (right) Geometric error under different texture decay levels with the normal dataset. The geometric error of the identity matrix is used as the fallback error.}
   \label{fig:experiment_results_analysis}
\end{figure}

Registration accuracy is measured by geometric error: the average distance between image corners transformed by the estimated and ground-truth homography. For failure cases, we use the identity matrix error as an upper bound for geometric error. We show the results of these experiments in \figurename~\ref{fig:experiment_results_analysis}. In the presence of strong linear structures, HOME shows the best performance, notably better than LSD/LBD. The curves of all methods other than HOME almost overlap with that of the identity matrix, indicating that they fail in most cases. HOME also performs best in weak-texture cases. While other methods cannot register images well under strong texture decay, HOME maintains sufficiently accurate registration performance even under severe texture decay, such as $\alpha=0.95$, where most texture appears to be erased. This synthetic image-pair experiment clearly supports the robustness of HOME against strong linear structures and weak textures, consistent with the previous experiment on video sequences.

\section{Conclusion}

HOME is a novel image-registration method for Hough-space matching. Using Hough-space extrema as keypoints and one-dimensional radial patterns as descriptors, HOME achieves robust registration in strongly linear structures with high computational efficiency. Although HOME is evaluated on homography estimation, its robustness to linear structures benefits line-based SLAM and VO applications, promising future directions. HOME-based homography estimation directly improves video stabilization and similar computer-vision tasks \cite{morikawa2021}. HOME also shows robustness to weak textures and can enhance performance in challenging scenarios such as low-light conditions.

Future work should investigate why HOME detects over 400 inlier matches even in weak-texture images, despite few prominent lines. Developing learning-based methods tailored to Hough-space matching characteristics is another promising direction. HOME's robustness to blur also warrants investigation. Since blur along the summation direction minimally affects the Hough transform, HOME may inherently resist blur.

Finally, while HOME demonstrates high accuracy and robustness in our experiments, it is not a universal solution. Its focus on video processing limits performance when registering images with large viewpoint changes. For example, when a robot moves forward rapidly, significant changes in scale alter the global Hough-space pattern, potentially causing registration failure. A multi-scale strategy may be a solution to this problem, but it is future work. Another issue is the aperture problem. Although HOME is inherently strong against this problem compared with image-space local methods, it still cannot handle images consisting of only purely parallel lines, because this is a mathematically unsolvable problem. In such cases, additional information, such as Inertial Measurement Unit (IMU) data, is required to resolve the ambiguity.

\section*{Acknowledgment}

The author is grateful to Takeshi Miura for fruitful discussions, and to Masaki Hilaga for his support and permission to undertake this research as part of corporate responsibilities. An AI language model was utilized to assist with English phrasing and the structural organization of this manuscript.

\bibliographystyle{IEEEtran}
\bibliography{references}

\end{document}